\documentclass[10pt,twocolumn,letterpaper]{article}

\usepackage{iccv}
\usepackage{times}
\usepackage{epsfig}
\usepackage{graphicx}
\usepackage{amsmath}
\usepackage{amssymb}
\usepackage{latexsym}

\usepackage{color}
\usepackage{subfigure}
\usepackage{bm}
\usepackage{algorithm}
\usepackage{algorithmic}
\usepackage{booktabs}

\usepackage{xcolor}

\newcommand*{\affaddr}[1]{#1} 
\newcommand*{\affmark}[1][*]{\textsuperscript{#1}}

\newcommand{\bsym}[1]{\boldsymbol #1}

\usepackage[pagebackref=true,breaklinks=true,colorlinks,bookmarks=false]{hyperref}

\iccvfinalcopy 

\ificcvfinal
\fi
\begin{document}

\title{Attract or Distract: Exploit the Margin of Open Set}

\author{
Qianyu Feng\affmark[1],
Guoliang Kang\affmark[2],
Hehe Fan\affmark[1,$3^*$],
Yi Yang\affmark[1]
\\
\affaddr{\affmark[1]
Centre for Artificial Intelligence, University of Technology Sydney
}\\
\affaddr{\affmark[2]
School of Computer Science, Carnegie Mellon University
}
\affaddr{\affmark[3]
Baidu Research
}\\
\tt\small{
\{qianyu.feng,hehe.fan\}@student.uts.edu.au, yi.yang@uts.edu.au,
kgl.prml@gmail.com}
}

\maketitle
\thispagestyle{empty}

\footnotetext[1]{Part of this work was done when Hehe Fan was an intern at Baidu Research.}%

\begin{abstract}

Open set domain adaptation aims to diminish the domain shift across domains, with partially shared classes.
There exist unknown target samples out of the knowledge of source domain.
Compared to the close set setting, how to separate the unknown (unshared) class from the known (shared) ones plays a key role.
Whereas, previous methods did not emphasize the semantic structure of the open set data, 
which may introduce bias into the domain alignment and 
confuse the classifier around the decision boundary.
In this paper, we exploit the semantic structure of open set data from two aspects:
1)  \textbf{Semantic Categorical Alignment}, 
which aims to achieve good separability of target known classes by categorically aligning the centroid of target with the source.
2)  \textbf{Semantic Contrastive Mapping},
which aims to push the unknown class away from the decision boundary.
Empirically, we demonstrate that our method performs favourably against the state-of-the-art methods on representative benchmarks, 
e.g. Digit datasets and Office-31 datasets.

\end{abstract}

\section{Introduction}
Recent days have witnessed the advancement in many computer vision tasks 
~\cite{krizhevsky2012imagenet,simonyan2014very,deeper2015,he2017mask,redmon2016you,long2015fully,he2016deep}.
The success achieved can be largely attributed to the sufficient amount of labeled in-domain data.
However, it is common that test data comes from a different distribution against the training data.
Such so-called domain shift may degenerate the model performance heavily.
Domain adaptation deals with this issue by diminishing the discrepancy across two domains.
The widely considered \textit{close set} setting assumes that both domains share the same set of underlying categories.
However, in practice, it is common that some unshared (unknown) classes exist in the target.
The methods developed for close set domain adaptation may not be trivially transferred to such \textit{open set} setting.

In this paper, we focus on the open set visual domain adaptation 
which aims to deal with the domain shift and the identification of unknown objects simultaneously,
in the absence of target domain labels.
Compared to the close set domain adaptation, how to separate the unknown class from the known ones plays a key role.
Up to now, the open set recognition still remains as a pending issue.
First raised by Busto \textit{et al.}~\cite{busto2017open}, they proposed to deal with open-set domain adaptation as an assignment task. 
\cite{mahsa2018facto} separated the unknown according to whether the sample can be reconstructed with the shared feature or not.
While the above methods use part of labeled data from uninteresting classes as unknown samples, it is not possible to represent all the unknown categories in the wild.
Another setting has been raised by Saito \textit{et al.}~\cite{saito2018eccv} where unknown samples only exist in the target domain, which is closer to a realistic scenario. 
Saito \textit{et al.} regarded the unknown samples as a separate class together with an adversarial loss to distinguish them.
It is worth noting that the existence of unknown samples hinders the alignment across domain. 
In the meanwhile, the disalignment inter-class across domain also makes it harder to distinguish the unknown samples.

\begin{figure}[t]
\centering
\includegraphics[width=\hsize]{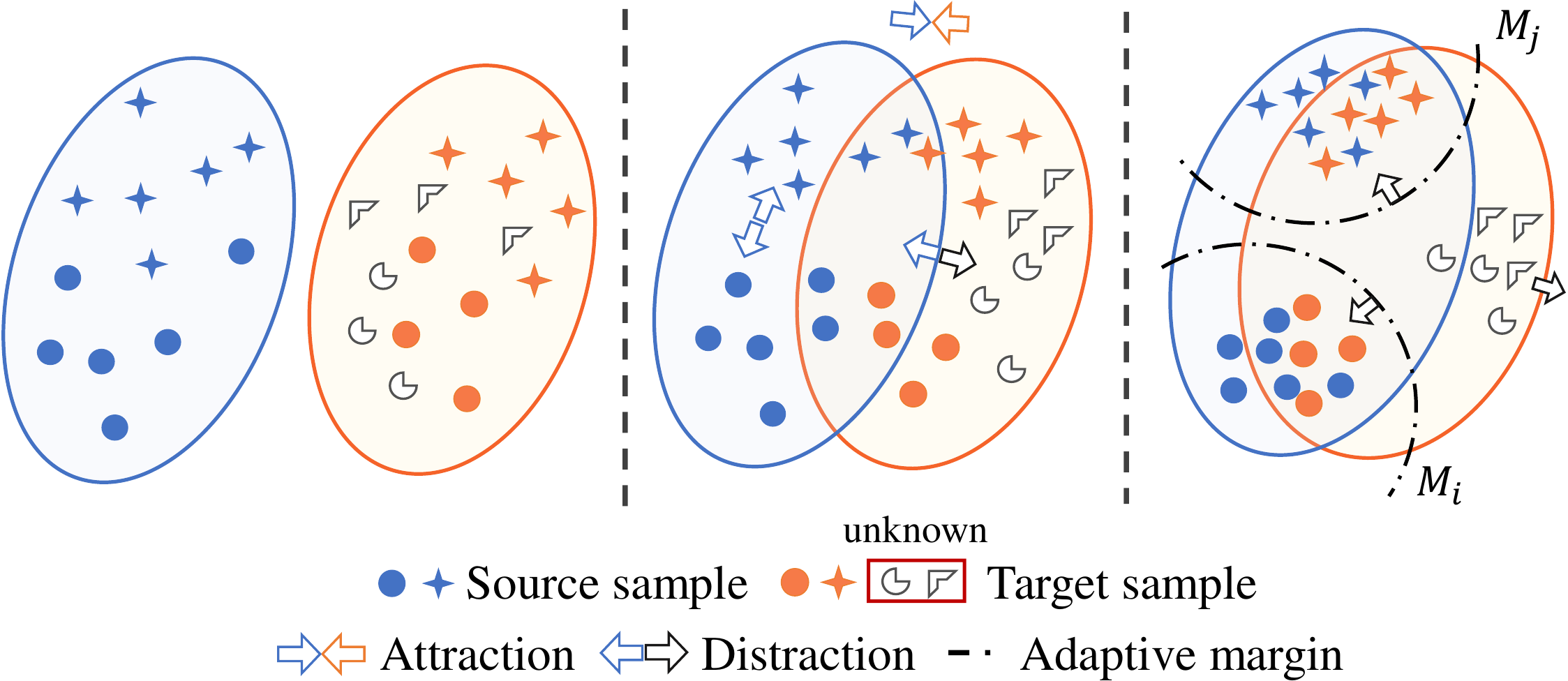}
\caption{Visualization of data distribution with the proposed method. {\bf Left}: Data before the adaptation with the existence of domain shift and unknown samples. {\bf Middle}: Neighbors from the same class are pulled closer, while samples from unknown class are pushed away. {\bf Right}: With the proposed method, representations become more discriminative. Samples from target domain can be better aligned within the corresponding neighborhood or distracted away from the known classes.}
\label{fig:intro}
\end{figure}

Considering the aforementioned problem, we take the semantic structure of open set data into account to 
make the unknown class more separable and thus
improve the model's predictive ability on target domain data.
Specifically, we focus on enlarging two kinds of margins: 
1) the margin across the known classes and 
2) the margin between the unknown samples and the known classes.
For the first one, we aim to make the known classes more separable 
and for the second one, we expect to push the unknown class away from the decision boundary.
As shown in Fig.~\ref{fig:intro}, 
during training, 
samples coming from different domains but within the same class (\emph{e.g.} the blue circle and the red circle) ``attract" each other.
For each domain, the margin between different known classes (\emph{e.g.} the blue circle and the blue star) are enlarged.
Moreover, the samples within unknown class (\emph{e.g.} the irregular polygons) are ``distracted" from the samples from known classes.

We propose using \textit{semantic categorical alignment (SCA)} and \textit{semantic contrastive mapping (SCM)} to achieve our goal.
For semantic categorical alignment, 
due to the absence of target annotations,
we indirectly promote the separability across target known classes 
by categorically aligning their centers with those in the source domain.
For the source domain,
although to an extent, the separability across known classes can be achieved 
through imposing the cross-entropy loss on the labeled data,
we explicitly model and enhance such separability with 
the contrastive-center loss \cite{QiS16eccv}.
Empirically we demonstrate that our method leads to more discriminative features and 
benefit the semantic alignment across domains.

Although the semantic categorical alignment helps make the decision boundary aligned across two domains, 
there may still exists confusing data of unknown class lying near the decision boundary.
Thus we propose using semantic contrastive mapping to push the unknown class away from the boundary.
In detail, we design the contrastive loss to make the margin 
between the unknown and known class larger than that between known classes.
 
As the target labels are not available,
we use the predictions of the network at each iteration as the hypothesis of target labels
to perform the semantic categorical alignment and the semantic contrastive mapping.
We start our training from the source trained model to give a good initialization of target label hypothesis.
Although the hypothesis of target labels may not be accurate, 
SCA and SCM in itself are robust to such noisy labels.
Empirically we find that the estimated SCA/SCM loss works as a good proxy
to improve the model's performance on the target domain data.

In a nutshell, our contributions can be summarized as 
\begin{itemize}
\item We propose using semantic categorical alignment to achieve  good  separability  of  target  known classes
and semantic contrastive mapping to push the unknown class away from the decision boundary.
Both benefits the adaptation performance noticeably.
\item Our method performs favourably against the-state-of-the-art methods on two 
representative benchmarks,
\emph{i.e.} on Digits dataset, it achieves 84.3\% accuracy on the average, 1.9\% higher than the state-of-the-art;
on Office-31 dataset, we achieve 89.7\% with AlexNet and 89.1\% with VGG.
\end{itemize}

\section{Related work}

Domain adaptation for visual recognition aims to bridge the knowledge gap between across different domains. 
Approaches for open set recognition attempt to figure out the unknown samples while identifying samples from known classes.
In a real scenario, data not only come from diverse domains but also varies in a wide range of categories.
This paper focuses on dealing with the overlap of these two problems.

Many methods~\cite{long2015learning,long2016unsupervised,ghifary2016deep,tzeng2017adv,bousmalis2017unsupervised,shu2018dirt,kang2018deep,zhong2019invariance,luo2019significance,zhong2019camstyle} have been proposed for unsupervised domain adaptation (UDA), including the deep network. 
These work bring significant results focusing on the \textit{closed set} in the following aspects. 
\textit{Distribution-based learning.} 
Many approaches aim to learn features invariant to domain with a distance metric~\cite{long2015learning,long2016unsupervised,tzeng2014deep}, 
\eg, KL-divergence, Maximum Mean Discrepancy (MMD), Wasserstein distance, but they neglected the alignment of conditional distribution.
The categorical information is exploited to align domains at a fine-grained level together with the pseudo-label.
The marginal distribution and conditional distribution can also be jointly aligned with a combined MMD proposed by~\cite{saito2017asymmetric}.
\cite{sener2016learning,haeusser2017associative,saito2017adversarial,saito2018maximum} pay attention to the discriminative property of the representation.
This paper is also related to work~\cite{kang2019contrastive, guan2018multi, chen2018virtual, luo2019taking,DBLP:conf/iccv/FanCCYXH17,DBLP:journals/tomccap/FanZYY18,zheng2017dual} considering the categorical semantic compactness and separability as the same time.
\textit{Task-oriented learning.} 
Approaches~\cite{tzeng2017adv, bousmalis2017unsupervised, ganin2014unsupervised} tend to align the domain discrepancy in an adversarial style.
Ganin \textit{et al.}~\cite{ganin2014unsupervised} proposed to learn domain-invariant feature by using an adversarial loss which reverses the gradients during the back-propagation.
Bousmalis \textit{et al.}~\cite{bousmalis2017unsupervised} enabled the network to separate the generated features into domain-specific subspace and domain-invariant subspace.

\begin{figure*}[t]
\centering
\includegraphics[width=0.92\hsize]{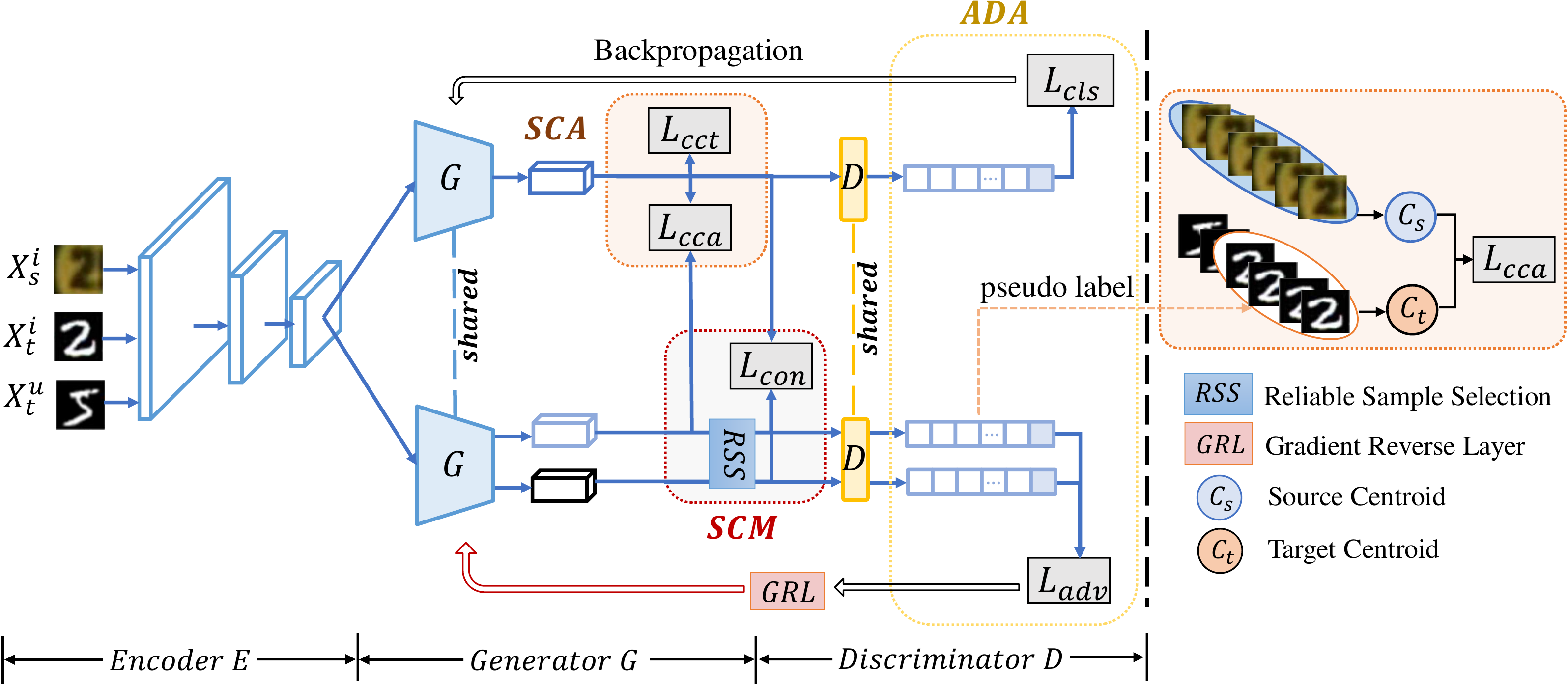}
\caption{Framework of the proposed method. There are three modules: Adversarial Domain Adaptation (ADA), Semantic Categorical Alignment (SCA) and Semantic Contrastive Mapping (SCM). SCA aims to learn discriminative representation and align samples from the same category across domains. SCM attempts to distract unknown samples away from all the known categories. All the modules are trained simultaneously and work together to better categorize each known class and unknown class.}
\label{fig:framework}
\end{figure*}

The aforementioned methods tackle with domain adaptation in the closed-set scenario. 
Inspired by recent work in open-set recognition, the problem of Open Set Domain Adaptation (OSDA) is raised by Busto \textit{et al.}~\cite{busto2017open}. 
With several classes trained as unknown samples, Busto \textit{et al.} proposed to solve this problem by learning a mapping across domains with an assignment problem of the target samples.
As is aforementioned, it is not possible to cover all the unknown samples with selected categories.
Another setting where the unknown class does not exist in the source domain is raised by Saito \textit{et al.}~\cite{saito2018eccv}.
Regarding unknown as a different class, they enable the network to align feature between known classes and reject the unknown samples at the same time.
\cite{mahsa2018facto} tried to separate unknown samples from known class by disentangling the representation into private and shared parts. 
They proved that the samples from the known classes can be reconstructed with shared features while the unknown samples can not.

Our method also regards the unknown samples as an ``unknown'' class. 
What is different, our method devotes to solve the open-set domain adaptation by enhancing the discriminative property of representation, aligning similar samples in the target with source domain while pushing the unknown samples away from all the known classes.

\section{Method}


\subsection{Overall Architecture}

The crucial problems of open set domain adaptation consist in two aspects, \ie, 
align the target known samples with the known samples in the source domain, 
and separate unknown samples in target from the target known samples.
To solve these two problems, we design the following modules.
\textbf{1) Adversarial Domain Adaptation (ADA).} 
Based on a cross-entropy loss, ADA aims to initially align samples in the target with source known samples or classify them as unknown. 
\textbf{2) Semantic Categorical Alignment (SCA).} 
This module consists of two parts.
First, based on a contrastive-center loss, aims to compact the representation of samples from the same class. 
Second, based on a center loss across domains, tries to align the distribution of the same class between source and target.
\textbf{3) Semantic Contrastive Mapping (SCM).} 
With a contrastive loss, SCM aims to encourage the known samples in the target to move closer to the corresponding centroid in source. 
While it also attempts to keep the unknown samples away from all the known classes.

The overall framework of our method is illustrated in Fig~\ref{fig:framework}. 
It consists of an encoder $E$, a generator $G$ and a discriminator $D$.
The image encoder $E$ is a pretrained CNN network to extract semantic features which may involve the domain variance. 
The feature generator $G$ is composed of a stack of fully-connected (FC) layers.
It aims to transform the image representation into a task-oriented feature space. 
The discriminator $D$ classifies each sample with the generated representation into a category.



\newcommand{\mymin}{\mathop{\rm min}\limits}
\newcommand{\mymax}{\mathop{\rm max}\limits}
\newcommand{\1}{\mbox{1}\hspace{-0.25em}\mbox{l}}

\subsection{Adversarial Domain Adaptation}

Suppose $\{X_{s},Y_{s}\}$ is a set of labeled images sampled from the source domain, in which each image $\bsym{x_{s}}$ is paired with a label $y_{s}$.
Another set of images $X_{t}$ derives from the target domain.
Different from $X_{s}$, each image $\bsym{x_{t}}$ in $X_{t}$ is unlabelled and may come from unknown classes. 
The goal of open set domain adaptation is to classify the input image $\bsym{x_{t}}$ into $N+1$ classes, where $N$ denotes the number of known classes. 
All samples from unknown classes are expected to be assigned to the unknown class $N+1$.

We leverage an adversarial training method to initially align samples in the target with source known samples or reject them as the unknown.
Specifically, the discriminator $D$ is trained to separate the source domain and the target domain.
However, the feature generator $G$ tries to minimize the difference between the source and the target.
When an expert $D$ fails to figure out which domain the sample comes from, the $G$ learns the domain-invariant representation.


We use the cross-entropy loss together with the softmax function for the known source samples classification,
\begin{equation}
\label{eq:loss_ce}
\begin{aligned}
\mathcal{L}_{cls}({x_s},y_s) &= -\log(p(y=y_s|{\bsym{x_s}})),\\
&= -\log(D \circ G({\bsym{x_s}}))_{y_s}).
\end{aligned}
\end{equation}
Following~\cite{saito2018eccv}, in an attempt to make a boundary for an unknown sample, we utilize a binary cross entropy loss,
\begin{equation}
\label{eq:loss_adv}
\begin{aligned}
\mathcal{L}_{adv}(\bsym{x_t}) = &- \frac{1}{2}\log(p(y=N+1|\bsym{x_t})) \\
  &- \frac{1}{2} \log(1-p(y=N+1|\bsym{x_t})).
\end{aligned}
\end{equation}

By the gradient reverse layer~\cite{ganin2014unsupervised}, we can flip the sign of the gradient during backward, which allows us to update the parameters of $G$ and $D$ simultaneously.
Then, the objective of the ADA module can be formulated as
\begin{eqnarray}
\label{eq:loss_ada}
\begin{aligned}
\mathcal{L}_{ADA} = &\mymin_{G} (\mathcal{L}_{cls}(\bsym{x_s},y_s) - \mathcal{L}_{adv}(\bsym{x_t})) + \\
                    &\mymin_{D} (\mathcal{L}_{cls}(\bsym{x_s},y_s) + \mathcal{L}_{adv}(\bsym{x_t})).
\end{aligned}
\end{eqnarray}

The ADA module only initially aligns samples in the target with source known samples and learns a rough boundary between the known and the unknown. 




\subsection{Semantic Categorical Alignment}
We try to address the issues existing in ADA by further exploring the semantic structure of open-set data.
To separate the unknown class from the known in the target domain, 
we should 1) make each known class more concentrate and the alignment between the source and the target more accurate
and 2) push the unknown class away from the decision boundary.
In this section, we aim to solve the first problem.
\\
We introduce the Semantic Categorical Alignment (SCA), which aims to compact the representation of known classes and distinguish each known class from others. 
There are two steps in SCA. 
First, the contrastive-center loss\cite{QiS16eccv} is adopted to enhance the discriminative property of generated features of source samples.
Second, each centroid of known classes from target will be aligned with the corresponding centroid of class in source domain. 
In this way, representations of source samples will finally become more discriminative, meanwhile, the known target centroids will be aligned more accurate.
\\
To compact the source samples that belong to the same class in the feature space, we apply the following contrastive-center loss to the source samples,
\begin{equation}\label{equ:contrastive-center loss}
\mathcal{L}_{cct}=\frac{1}{2}\sum_{i=1}^m\frac{ \|x_s^i-c_s^{y_s^i}\|_2^2}{(\sum_{j=1,j\neq{y_s^i}}^N\|x_s^i-c_s^{j}\|_2^2) +\delta},
\end{equation}
where $m$ denotes the number of samples in a mini-batch during training procedure, $x_s^i$  denotes the $i$-th training sample from the source domain. 
$c_s^{y_s^i}$ denotes the centroid of class $y_s^i$ in the source domain. 
$\delta$ is a constant used for preventing zero-denominator. In our experiments, $\delta=1$ is set to be $10^{-6}$ by default.


To align the two centroids of a known class between the source and target, we try to minimize the distance between the pair of centroids $dist(c_s^k, c_t^k)=\left\Vert{c_s^k-c_t^k}\right\Vert^2$, 
where $c_s^k$ and $c_t^k$ represent the centroids of class $k$ from the source and target domain, respectively. 

Due to the randomness and deviation in each mini-batch, we align the global centroids (calculated from all samples) instead of the local centroids (calculated from a mini-batch). 
However, it is not easy to directly obtain the global centroids.
We propose to partially update them with the local centroids at every iteration, according to their cosine similarities to the centroids in the source domain.
Specifically, we first compute the initial global centroids based on the prediction of the pretrained model as follows,
\begin{equation}\label{eq7}
c_{(0)}^k=\frac{1}{n^k}\sum_{j=0}^{n^k}G(x_i^{k}),
\end{equation}
where $n^k$ denotes the number of samples with label as $k$.
The pretrained model is trained on the source domain within the supervised classification paradigm.
For the target samples, we use the results of prediction as pseudo labels. 
In each iteration, we compute a set of local centroids $a_{(I)}^k$ using the mini-batch samples, where $I$ denotes the iteration. 
We compute the local centroids as the average of all samples in each iteration.
Then, the source centroid $c_s^k$ and target centroid $c_t^k$ are updated with re-weighting as follows,
\begin{equation}\label{eq9-1}
\begin{split}
\rho_s&=\rho(a_{s(I)}^k, c_{s(I-1)}^k),\\
c_{s(I)}^k&\leftarrow{\rho_s{a_{s(I)}^k}+(1-\rho_s)c_{s(I-1)}^k},
\end{split}
\end{equation}
\begin{equation}\label{eq9-2}
\begin{split}
\rho_t&=\rho(a_{t(I)}^k, c_{s(I-1)}^k),\\
c_{t(I)}^k&\leftarrow{\rho_t{a_{t(I)}^k}+(1-\rho_t)c_{t(I-1)}^k},
\end{split}
\end{equation}
where $\rho(\cdot,\cdot)$ is defined as 
$\rho(x_i,x_j) = (\frac{x_i \cdot x_j}{\|x_i\| \times \|x_j\|} + 1) / 2.$
Finally, the categorical center alignment loss is formulated as follows,
\begin{equation}\label{eq19}
\mathcal{L}_{cca}=\sum\limits_{k=1}^{N}dist(c_{s(I)}^k, c_{t(I)}^k).
\end{equation}


The benefits of SCA are intuitive:
1) The contrastive center loss, \ie, Eq.~(\ref{equ:contrastive-center loss}), enhances the compactness of the representations which also enlarges the margin inter-class.
2) The categorical center alignment loss, Eq.~(\ref{eq19}), guarantees that 
the centroids of the same class are aligned between the source domain and the target domain.
3) The dynamic update together ensures that the SCA aligns the global and up-to-date categorical distributions. 
Furthermore, the reweighting technique weakens the incorrect pseudo-labels and therefore can alleviate the accumulated error of the pseudo-labels.

\subsection{Semantic Contrastive Mapping}

SCA aligns the centroids of the same class between the source domain and the target domain.
For the non-centroid samples in the target domain, we employ a contrastive loss function to encourage the known samples to move closer to their centroids and enforce the unknown samples to stay far away from all the centroids of known classes.
By this way, we can align the non-centroid samples in the target domain.
We refer to this process as the Semantic Contrastive Mapping (SCM).

Since the pseudo labels of target samples are not totally correct, we select reliable samples whose classification probabilities are over a threshold.
We set the threshold to $\frac{1}{N+1}$ in our method.
SCM aims to reduce the distance between the reliable known samples and their centroids, while enlarge the distance between the reliable unknown samples and all centroids. 
\begin{equation}\label{con_loss}
\mathcal{L}_{con}(x_t;G) = (1-z)\mathcal{D}_{k}(x_t^k, c_s^k) - \frac{z}{N} \sum_{k=1}^{N}\mathcal{D}_{u}(x_t^k, c_s^k),
\end{equation}
where $z$ is equal to 0 if $x_t$ is predicted as class $\in{1,2,...,N}$, otherwise, z equals to 1. 
$\mathcal{D}_{k}$ denotes the distance between target known samples and the corresponding source centroid.
$\mathcal{D}_{u}$ denotes the distance between target unknown samples and all the source known classes.
Inspired by the energy-base model in~\cite{lecun2006cvpr}, functions are designed as follows,
\begin{equation}\label{sim_loss}
\mathcal{D}_{k}(x_t^k, c_s^k) = (1-\rho)^\omega dist(x_t^k, c_s^k)^2,
\end{equation}
\begin{equation}\label{dis_loss}
\mathcal{D}_{u}(x_t^{N+1}, c_s^k) = - \rho^\omega(max\{0, M^k - dist(x_t^{N+1}, c_s^k)\})^2,
\end{equation}
where $\rho$ denotes the cosine similarity. 
To ensure an efficient and accurate measurement of the distances, we also use a hyper-parameter $\omega$ to re-weight distances calculated in the loss. 
$M^k$ is a categorical adaptive margin to measure the radius of neighborhood of class $k$, defined as follows, 
\begin{equation}\label{eq:mk}
M^k = \frac{1}{N}\sum_{j=1,j\neq{k}}^{N} dist(c_t^j, c_s^k).
\end{equation}


\subsection{Objective}

\begin{algorithm}[!t]
\caption{Exploit the Margin of Open Set, e denotes the training step, $I$ denotes the iteration times.}\label{algorithm1}
  \begin{algorithmic}[1]
  \REQUIRE Labeled samples batches $X_s=\{(x_{s_i},y_{s_i})\}_{i=1}^{n_s}$ from source domain, unlabeled samples batches $X_t=\{x_{t_j}\}_{j=1}^{n_t}$ from target domain.
  $B_i^s$ and $B_i^t$ denote the $i_{th}$ mini-batch data in the training set.
  \ENSURE Parameters in the network $\theta_G,\theta_D$
  \STATE \textbf{\emph{1st Stage}}
  \STATE Pretrain $G$ and $D$ based on $\{X_s, Y_s\}$, update $\theta_G,\theta_D$
  
  \STATE \textbf{\emph{2nd Stage}}
  \STATE $e=0$
  \WHILE {not converge}
  \STATE Calculate the current global centroids $c_{s(e)}^k$ and $c_{t(e)}^k$ 
  \FOR {$I=1$ \textbf{to} $max\_iter$}
  \STATE Update $c_{s(e)}^k$ and $c_{t(e)}^k$ by using Eq. \ref{eq9-1} and Eq. \ref{eq9-2}
  \STATE Calculate pair distance between $c_{s(e)}^k$ and $c_{t(e)}^k$
  \STATE Select reliable target samples $\hat{X}_{t(I)}$
  \STATE Calculate pair distance between $c_{s(e)}^k$, $\hat{X}_{t(I)}$
  \STATE Train $model_m$ with $B_i^s, B_i^t$ by optimizing loss in Eq.~\ref{equ:total_loss}, update $\theta_G,\theta_D$
  
  \STATE $e=e+1$
  \ENDFOR
  \ENDWHILE
  \end{algorithmic}
\end{algorithm}

In the proposed method, considering the intra-class compactness and inter-class separability, 
we design the two modules SCA and SCM based on the adversarial learning in ADA. 
Formally, the final objective is defined in Eq.~\ref{equ:total_loss}. 
\begin{equation}\label{equ:total_loss}
\begin{aligned}
\mathcal{L}_{total} &= \mathcal{L}_{ADA} + \mathcal{L}_{SCA} + \mathcal{L}_{SCM}\\
                    &= \mathcal{L}_{cls} + \mathcal{L}_{adv} + \lambda_s \mathcal{L}_{cct} + \lambda_{c} \mathcal{L}_{cca} + \lambda_t \mathcal{L}_{con}.
\end{aligned}
\end{equation}

In each iteration, the network updates the class centroids and network parameters simultaneously. 
The overall algorithm is shown in Algorithm~\ref{algorithm1}.
SCA attempts to enlarge the margins between known classes in source and categorically align the centroids across domain.
SCM attempts to align all the known target samples to its source neighborhoods,
while keeping the distance between unknown samples and the centroids of known classes
around an adaptively determined margin.
With SCA, the discriminator in ADA is access to more discriminative representation and well-aligned semantic features.
On the other side, SCM aids to distinguish the unknown samples from the other known classes.

\section{Experiments}

\begin{table*}[t]
\centering
\scalebox{0.8}{
\begin{tabular}{c||cccc|cccc|cccc||cccc}
 \toprule[1.5pt]
& \multicolumn{4}{|c}{SVHN-MNIST}   &  \multicolumn{4}{|c|}{USPS-MNIST} & \multicolumn{4}{|c||}{MNIST-USPS}&\multicolumn{4}{|c}{Average}\\
  \multicolumn{1}{c||}{Method}    & OS & OS* & ALL & UNK  &  OS & OS* & ALL & UNK  
  & OS & OS* & ALL & UNK  &  OS & OS* & ALL & UNK\\\hline

OSVM &54.3&63.1&37.4&10.5&43.1&32.3&63.5&97.5&79.8&77.9&84.2&89.0&59.1&57.7&61.7&65.7\\
MMD+OSVM &55.9& 64.7&39.1&12.2&62.8&58.9&69.5&82.1&80.0&79.8&81.3&81.0&68.0&68.8&66.3&58.4\\	
BP+OSVM &62.9&{\bf 75.3}&39.2&0.7&84.4& 92.4&72.9&0.9&33.8&40.5&21.4&44.3&60.4&69.4&44.5&15.3\\
OSDA+BP\cite{saito2018eccv} & 63.0&59.1&71.0 &82.3&92.3&91.2&{\bf 94.4}&{\bf97.6}&{\bf 92.1}&{\bf 94.9}&88.1&78.0&82.4&81.7&84.5&85.9\\
\midrule	
Ours w/o SCA &65.6& 61.6&73.9&85.4 &93.6&95.4&87.9&82.8 &86.5&86.1&88.1&{\bf 88.5} &81.9&81.0&83.3&85.6\\
Ours w/o SCM &65.5& 61.0&74.8&87.8 &92.5&93.8&87.1&81.1 &84.6&84.0&86.0&87.7 &80.9&79.6&82.6&85.5\\\hline
Ours & {\bf 68.6}&65.5&{\bf 75.3}&{\bf 84.3}
     & {\bf 93.1}&{\bf95.2}&92.8& 91.7
     & 91.3&92.0&{\bf 90.7}&87.8
     & {\bf 84.3}&{\bf 84.2}&{\bf 86.3}&{\bf 87.9}\\
\bottomrule[1.5pt]
\end{tabular}}
\vspace{2mm}
\caption{Accuracy (\%) of experiments on Digit dataset.}
\label{tab:digits}
\end{table*}

\begin{figure*}[t]
\centering
\includegraphics[width=0.92\hsize]{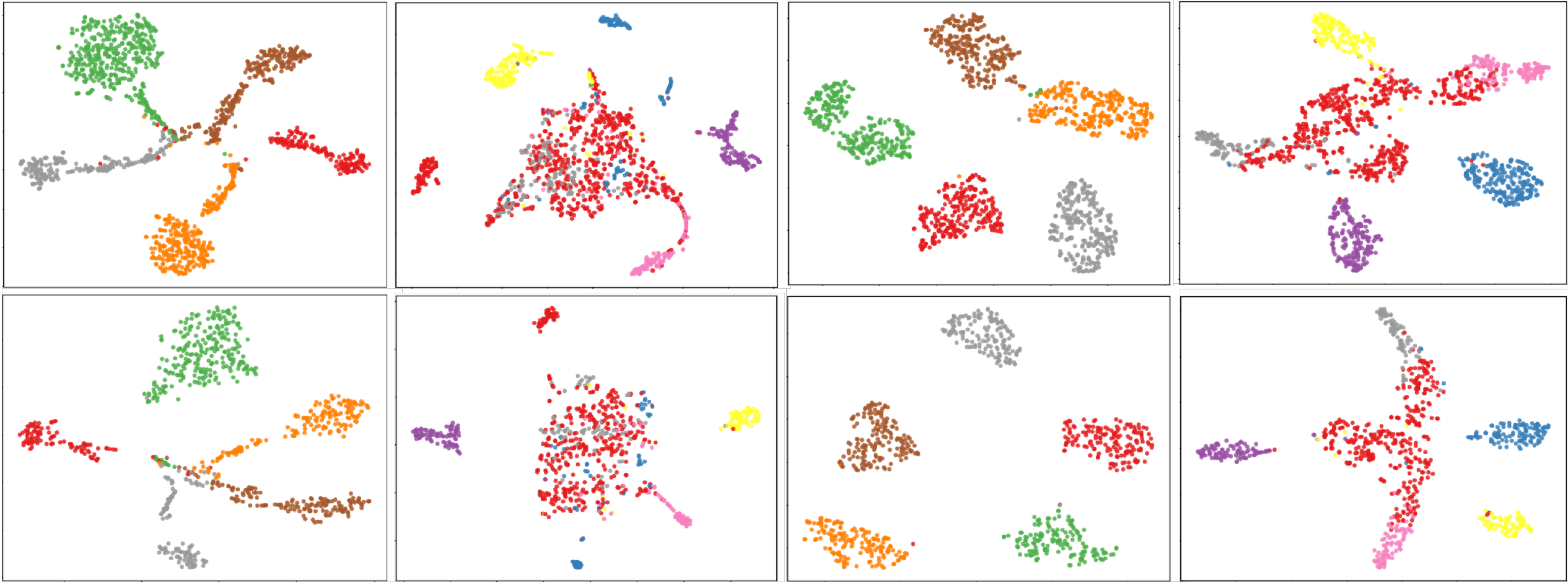}
\caption{A comparison between the existing method and the proposed method. 
\textbf{First row}: visualization of features of the state-of-the-art method OSDA+BP\cite{saito2018eccv}~. 
\textbf{Second row}: visualization of features generated by our method. 
The left two column show features of source and target in SVHN $\rightarrow$ MNIST, the right two columns are features of source and target in MNIST $\rightarrow$ USPS. The color red in the target represents the unknown class.}
\label{fig:digit-tsne}
\end{figure*}

\subsection{Setup}
In this section, we evaluate the proposed method on the open set domain adaptation task using two benchmarks, \ie, Digit datasets  and \textit{Office-31}~\cite{saenko2010adapting}. 
Considering the setting where unknown samples only exist in the target domain. 
We compare the performance of our method OSDA+BP~\cite{saito2018eccv} and other baselines including: 
Open-set SVM (OSVM)~\cite{jain2014multi} and other methods combined with OSVM, \eg, 
Maximum Mean Discrepancy(MMD)~\cite{gretton2007kernel}, BP~\cite{ganin2014unsupervised}, ATI-$\lambda$~\cite{busto2017open}.
OSVM classifies test samples into unknown class with a threshold of probability when the predicted probability is lower than the threshold for other classes. 
OSVM also requires no unknown samples in the source domain during training. 
MMD+OSVM is a combination method with OSVM and MMD-based method for network in~\cite{long2015learning}. 
MMD is discrepancy measure metric used to match the distribution across domains. 
BP+OSVM combines OSVM with a domain classifier, BP~\cite{ganin2014unsupervised}, which is a representative of adversarial learning applied in unsupervised domain adaptation.


\textbf{Digits} We begin by exploring three Digit datasets, \emph{i.e.} SVHN~\cite{netzer2011reading}, 
MNIST~\cite{lecun1998gradient} and USPS~\cite{lecun1998gradient}. 
SVHN contains colored digit images of size $32\times32$, where more than one digit may appear in a single image.
MNIST includes $28\times28$ grey digit images and USPS consists of $16\times16$ grey digit images.
We conduct 3 common adaptation scenarios including SVHN to MNIST, USPS to MNIST and MNIST to USPS. 

\textbf{Office-31}~\cite{saenko2010adapting} is a standard benchmark for domain adaptation. 
There exist three distinct domains: Amazon (A) with 2817 images from the merchants, Webcam (W) with 795 images of low resolution and DSLR (D) with 498 images of high resolution. 
Each domain shares 31 categories with the others. 
We examine the full transfer scenarios in our experiments.

\textbf{Implementation} 
For Digit datasets, we employ the same architecture with \cite{saito2018eccv}.
For \textit{Office-31}, we employ two representative CNN architectures, 
AlexNet~\cite{krizhevsky2012imagenet} and VGGNet~\cite{simonyan2014very}, to extract the visual features. 
For both the generator and classifier, we use one-layer FC followed with Leaky-RELU and Batch-Normalization.
For \textit{Office-31}, 
we initialize the feature extractor from the ImageNet~\cite{deng2009imagenet} pretrained model
For both datasets, we first train our model with labeled source domain data.
All networks are trained by Adam \cite{kingma2014adam} optimizer with weight decay $10^{-6}$. 
The initial learning rates for Digit and \textit{Office-31} datasets is $2\times10^{-4}$ 
and $10^{-3}$ respectively.
Learning rate decreases following a cosine ramp-down schedule.
%
We set the hyper-parameters $\lambda_s=0.02$, $\lambda_{c}=0.005$,  and $\lambda_t=10^{-4}$ in all the experiments.
Following \cite{busto2017open}, we report the accuracy averaged over the classes in the OS and OS*. 
The average accuracy of all classes including the unknown one is denoted as OS. 
Accuracy measures only on the known classes of the target domain is denoted as OS*.
All the results reported are the accuracy averaged over three independent running.

\begin{table*}[t]
\centering
\scalebox{0.9}{
\begin{tabular}{c||cc|cc|cc|cc|cc|cc|cc}
\cline{2-15}
 \multicolumn{1}{c}{}  & \multicolumn{14}{|c|}{{\bf Adaptation Scenario}} \\
\cline{2-15}
  \multicolumn{1}{c}{}  & \multicolumn{2}{|c}{A-D}   &  \multicolumn{2}{|c|}{A-W} & \multicolumn{2}{c}{D-A}& \multicolumn{2}{|c|}{D-W}&\multicolumn{2}{c}{W-A}&\multicolumn{2}{|c|}{W-D}&\multicolumn{2}{c|}{AVG} \\
  \multicolumn{1}{c|}{}    &   OS   & OS*    & OS   & OS*  & OS   & OS*   & OS  & OS*& OS  & OS*& OS  & OS*& OS  &  \multicolumn{1}{c|}{OS*} \\\cline{2-7}\cline{2-15}
  \toprule[1.0pt]
\multicolumn{11}{c}{{\bf Method w/o unknown classes in source domain (AlexNet)}}&\multicolumn{4}{c}{}\\\hline
OSVM &59.6&59.1&57.1&55.0&14.3&5.9&44.1&39.3&13.0&4.5&62.5&59.2&40.6&37.1\\
MMD + OSVM&47.8&44.3&41.5&36.2&9.9&0.9&34.4&28.4&11.5&2.7&62.0&58.5&34.5&28.5\\
BP+OSVM&40.8&35.6&31.0&24.3&10.4&1.5&33.6&27.3&11.5&2.7&49.7&44.8&29.5&22.7\\
ATI-$\lambda$\cite{busto2017open} + OSVM&72.0&-&65.3&-&66.4&-&82.2&-&71.6&-&92.7&-&75.0&-\\
OSDA+BP\cite{saito2018eccv} &  76.6 & 76.4 & 70.1 & 69.1 & 62.5 & 62.3 & 94.4 & 94.6 & {\bf 82.3} & {\bf 82.2} & 96.8 & 96.9 & 80.4 & 80.2\\\hline
Ours w/o SCA &87.8&89.0&85.6&87.1&74.2&73.8&97.2&98.1&74.9&73.9&98.5&99.0&86.5&87.0\\
Ours w/o SCM &89.8&91.2&88.0&{\bf 90.6}&77.8&77.9&97.6&98.6&75.1&75.0&98.0&99.3&87.7&88.8\\\hline
Ours &  {\bf 91.0} & {\bf 92.7} & {\bf 89.5} & 89.6 & {\bf81.8} & {\bf 83.0} & {\bf 97.8} & {\bf 98.8} & 78.7 & 81.4 & {\bf 98.5} & {\bf 99.7} & {\bf 89.7} & {\bf 90.7}\\\hline

 \multicolumn{11}{c}{{\bf Method w/o unknown classes in source domain (VGGNet)}}&\multicolumn{4}{c}{}\\\hline
OSVM &82.1&83.9&75.9&75.8&38.0&33.1&57.8&54.4&54.5&50.7&83.6&83.3&65.3&63.5\\
MMD + OSVM&84.4&85.8&75.6&75.7&41.3&35.9&61.9&58.7&50.1&45.6&84.3&83.4&66.3&64.2\\
BP+OSVM&83.1&84.7&76.3&76.1&41.6&36.5&61.1&57.7&53.7&49.9&82.9&82.0&66.4&64.5\\
OSDA+BP\cite{saito2018eccv}  & 85.8 & 85.8 & 76.9 & 76.6 & {\bf 89.4} & {\bf91.5} & 96.0 & 96.6 & {\bf 83.4} & {\bf 83.1} & 97.1 & 97.3 & 88.0 & 88.5\\\hline
Ours  & {\bf90.1} & {\bf 92.0} & {\bf 86.4} & {\bf 87.7}& 81.6 & 88.4 &{\bf 97.9} &{\bf 99.8}&80.3&82.6&{\bf 98.2}&{\bf99.3}&{\bf 89.1}&{\bf 91.6}\\
\bottomrule[1.0pt]
\end{tabular}}
\vspace{0.5mm}
\caption{Accuracy (\%) of each method on scores of OS and OS*. 
A, D and W correspond to Amazon, DSLR and Webcam respectively. 
The ablation versions of our method w/o SCA and w/o SCM are also reported.}
\vspace{-2mm}
\label{tab:office-10}
\end{table*}

\subsection{Results on Digit Dataset}
In the three Digit datasets, numbers 0$\sim$4 are chosen as known classes. 
Samples from the known classes make up the source samples. 
In the target samples, numbers 5$\sim$9 are regarded as one unknown class.
Besides the scores of OS and OS*, we also report the total accuracy for samples in target and the accuracy of unknown class, which are denoted as \textit{ALL} and \textit{UNK}, respectively. 

As shown in Table~\ref{tab:digits}, our method produces competitive results compared to other methods. Results of our method outperform the other methods in SVHN $\rightarrow$ MNIST, MNIST $\rightarrow$ USPS and the average scores. 
It is shown that our method is better at recognizing the unknown samples(5$\sim$9) while maintaining the performance of identifying the known classes. 
For SVHN $\rightarrow$ MNIST, the semantic gap between them is large, as there may exist several digits in the images of SVHN. 
Thus the accuracies of SVHN $\rightarrow$ MNIST are lower than the other two scenarios.
Our method outperforms existing methods on the average scores.
For instance, our approach achieves 87.9\% in the average score of unknown class. This is 2.0\% higher than OSDA+BP \cite{saito2018eccv}.
Learned features obtained by the trained model of the three scenarios are visualized in Fig~\ref{fig:digit-tsne}. 
We observe that the distribution of unknown samples is more decentralized in SVHN $\rightarrow$ MNIST because of the large divergence of the two domains.
Compared with OSDA+BP \cite{saito2018eccv}, the proposed method could better centralize samples of the same known class and distinguish unknown samples from known ones.

\subsection{Results on Office-31 Dataset}

We compare our method with other works on \textit{Office-31} dataset following \cite{busto2017open}. 
There are 31 classes in this dataset, and the first 10 classes in alphabetical order are selected as known classes. 
In ~\cite{saito2018eccv}, 21-31 classes are selected as unknown samples in the target domain, which only exist in the target domain.
We evaluate the experiment on all the 6 scenario tasks: A $\rightarrow$ D, A $\rightarrow$ W, D $\rightarrow$ A, D $\rightarrow$ W, W $\rightarrow$ A, W $\rightarrow$ D.
The improvement on some hard transfer tasks is encouraging to prove the effectiveness and value of the proposed method.

Results of \textit{Office-31} are shown in Table~\ref{tab:office-10}. 
With features extracted with AlexNet, our method significantly outperforms the state-of-the-art methods except W $\rightarrow$ A.
Our approach achieves 89.7\% in OS score based on AlexNet, overpassing the state-of-the-art by 2\%. For OS*, our approach improves the results of state-of-the-art from 80.2\% to 90.7\%.
Based on VGG, our method reaches 89.1\% on OS score and 91.6 on OS* score, outperforming the state-of-the-art method by 1.1\% and 3.1\%, respectively.
Especially, under the harder scenarios of A $\rightarrow$ D, A $\rightarrow$ W, D $\rightarrow$ A, our method brings large improvement.

\subsection{Ablation Study}
\begin{figure*}[t]
\centering
\includegraphics[width=\hsize]{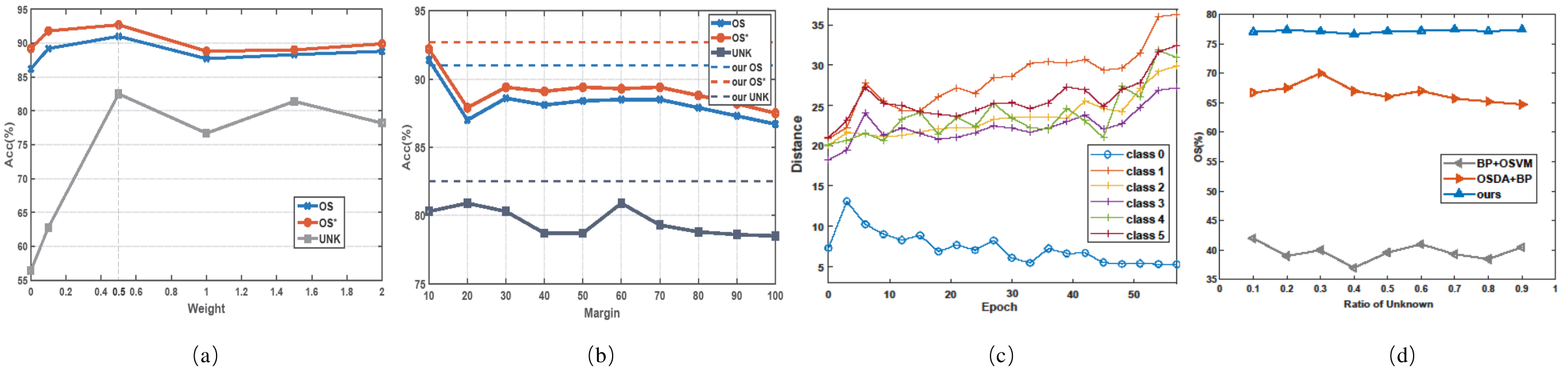}
\caption{\textbf{(a)}: A comparison of the behavior of our method with different re-weight $\omega$ in the contrastive loss.
\textbf{(b)}: A comparison between static margin and our adaptive margin.
\textbf{(c)}: Distances between centroid of class ``backpack'' labeled ``0'' in the target with centroids of class 0$\sim$5 in source .
\textbf{(d)}: Performance of the proposed method under different ratios of unknown samples.}
\label{fig:ablation}
\vspace{-1mm}
\end{figure*}
For a straightforward understanding of the proposed method, we further evaluate each module via ablation experiments on Digit datasets.
We alternately remove the SCA and SCM from our model.
Results are reported in Table~\ref{tab:digits} above our final results.
A decrease in performance is observed when removing SCA or SCM. 
Particularly, when the discriminative learning (w/o SCA) or contrastive learning (w/o SCM) is ablated, the accuracy of OS* or UNK or both of them will decrease significantly.
We reconfirm the effect of each module based on AlexNet in the experiments of Office-31. Results in~Table~\ref{tab:office-10} also indicate the importance of learning the discriminative representation and contrastive mapping simultaneously.
It indicates that the two modules take effect jointly.
The discriminative representation helps to push away the unknown samples while the distraction of unknown samples assists the alignment of known categories.
\\
\textbf{Effect of adaptive margin.}
As the margin is designed for the contrastive loss as an adaptive one, we also observe the behavior of the model with static margin on A $\rightarrow$ D.
We choose constant value of margin $m_s \in \{10, 20, 30, ..., 100\}$ for comparison.
When $m_s$ is equal to 0, the contrastive term in the objective is only to align all the target samples predicted as known with the corresponding centroid in source.
When $m_s$ is assigned with a large value, the model tends to penalize all the target samples predicted as unknown samples with large loss.
According to results in Fig.~\ref{fig:ablation}(b), the accuracies of OS and OS* are trending downward when using a constant margin.
The accuracy of UNK raises when the margin is 20 and 60. 
To further investigate the changes of distances between categories during the training of model,
we visualize the centroid of known class ``Backpack'' in the target with the centroids of known classes in the source domain. The ``Backpack'' is labeled as 0. Results are shown in Fig.~\ref{fig:ablation}(c).
With the training of model, the distance between the centroids of class 0 in the source and target domains is declining.
This indicates that the distribution between the source and target domains are aligned for class 0. 
In the meanwhile, distances between centroids of target class 0 and the other classes in source domain improve with the increase of training epoch. 
In spite of the reduce of discrepancy between the two domains, the distances with the other classes in source are increasing.
This demonstrates that it is improper to use a static margin to measure the energy for pulling apart unknown samples. 
With the alignment across domain and the separation between different class, the radius of the neighborhood of each class would change.
This also explains the reason why the adaptive margin produces higher scores than the static margin.
\\
\textbf{Effect of re-weighting the contrastive loss.}
There is another hyper-parameter $\omega$ which re-weights the distances in $\mathcal{L}_{con}$. 
We conduct experiments with $\omega \in \{0, 0.1, 0.5, 1, 1.5, 2\}$. As shown in Fig.~\ref{fig:ablation}(a), our approach achieves the best results when $\omega$ is equal to 0.5. 
This parameter smooths the weight of calculated with cosine similarity.
When $\omega$ is 0, the re-weighting term is equal to 1, which means the contrastive loss is calculated without the re-weighting. 
It also reveals that the effectiveness of the re-weighting term could help the model to better measure the distances between unknown samples and centroids in source.

\textbf{Effect of the ratio of unknown samples.}
In this section, we aim to investigate the robustness of our method under different ratios of unknown samples in the target data.
In Fig.~\ref{fig:ablation}(d), we compare our results with method BP+OSVM and OSDA+BP under ratio $\in (0, 1)$.  
Unknown samples are randomly sampled according to the ratio. 
It can be seen that the accuracy of our model fluctuates little and is always above the baseline methods, 
which implies the robustness of our method.

\section{Conclusion}

In this paper, we focus on the open set domain adaptation setting where 
only source domain annotations are available and 
the unknown class exists in the target. 
To better identify the unknown and meanwhile diminish the domain shift,
we take the semantic margin of open set data into account through
semantic categorical alignment and semantic contrastive mapping,
aiming to make the known classes more separable and
push the unknown class away from the decision boundary, respectively.
Empirically, we demonstrate that our method is comparable to or even better than the state-of-the-art methods
on representative open set benchmarks, \emph{i.e.} Digits and Office-31.
The effectiveness of each component of our method is also verified.
Our method implies that explicitly taking the semantic margin of open set data into account is beneficial.
And it is promising to make more exploration in this direction in the future.

{\small
\bibliographystyle{ieee}
\bibliography{main}
}

\end{document}